\documentclass[runningheads]{llncs}

\usepackage[T1]{fontenc}
\usepackage{graphicx}
\usepackage{optidef}
\usepackage{subcaption}
\usepackage{overpic}

\usepackage{amssymb}

\usepackage[citestyle=numeric, uniquelist = false, maxcitenames = 1, sorting=none]{biblatex}

\bibliography{ref}

\begin{document}

\title{Multi-Agent Deep Reinforcement Learning for Multi Objective Battery Management in Dairy Farms}

\titlerunning{MORL for Battery Management}

\author{Marcos Eduardo Cruz Victorio\orcidID{0000-0003-2604-176X} \and
Karl Mason\orcidID{0000-0002-8966-9100}}

\authorrunning{M. E. Cruz Victorio et al.}

\institute{University of Galway, Ireland\\
\email{marcos.cruzvictorio@universityofgalway.ie}
}

\maketitle  
\begin{abstract}
The dairy industry in Ireland has a large potential for the integration of renewable energy and the reduction of carbon emissions. However, researchers of distributed generation control are mainly focused on residential and commercial applications. To contribute to the effective integration of renewable energy in the dairy sector, this paper presents a multi-objective optimisation control system based on differential evolution and multi agent Deep Reinforcement Learning. The proposed control is organised in two layers: the upper layer uses dynamic pricing, and the lower layer is based on multi-agent reinforcement learning for battery management.
This paper also simulates the electrical response of the proposed control system in a rural distribution circuit. The simulation results show that the proposed control framework can improve profits from energy arbitrage up to 18\% compared to using Rule-based models, increase the use of distributed generation without significantly increasing cost, and comply with the Irish grid code in terms of voltage variation. 

\keywords{Multi-Objective Optimisation \and Deep Reinforcement \\Learning \and Distributed Generation.}
\end{abstract}

\section{Introduction}

Currently, there is a large focus on research to deliver the energy transition to combat climate change. In the case of electricity, research on the energy transition focuses on the integration of Distributed Energy Resources (DER)s. In this context, there is a large potential in the agricultural sector for the integration of DERs, due to its energy intensive nature and limited use of smart grid technologies. 

In an electrical grid, there are multiple, often competing, objectives that need to be balanced in the smart grid \cite {r14}, such as grid compliance, minimisation of cost and minimisation of carbon emissions. Although previous approaches have been proposed for grid multi objective energy optimisation (MOO) \cite{r15,r16}, their performance is limited by the precision of their models. In this context, Deep Reinforcement Learning (DRL) methods can be used for energy optimisation in conditions with limited information of the system dynamics. The applications of DRL in smart grids include maximising the use of renewable energy \cite{r3} and minimising costs in electric vehicles \cite{r28}. Among DRL methods, Proximal Policy Optimisation (PPO) emerges as an ideal method for energy management \cite{r36}.

MMO applications in smart grid have been presented with some limiations \cite{r32,r37}. For example, in \cite{r19} MOO is applied to reduce line congestion and costs, however, this approach did not account for multiple generators and storages in the grid. In \cite{r20}, a MOO framework is proposed for cost minimisation and voltage regulation, however, this approach is limited to the retail electricity market.

To further develop smart grid applications in the rural sector, our proposed control framework combines heuristic optimisation with DRL agents to maximise the use of distributed generation and affordability in dairy farms, accounting for uncertain electricity prices and multiple generation and storage sources. The following section describes the proposed energy management framework.

\section{Methodology}

\begin{figure}
\centering
\begin{minipage}[c]{0.5\textwidth}
\centering
\begin{subfigure}{1\textwidth}
    \begin{overpic}[width=1\linewidth, trim = {0.0cm 10cm 23cm 0cm}, clip]{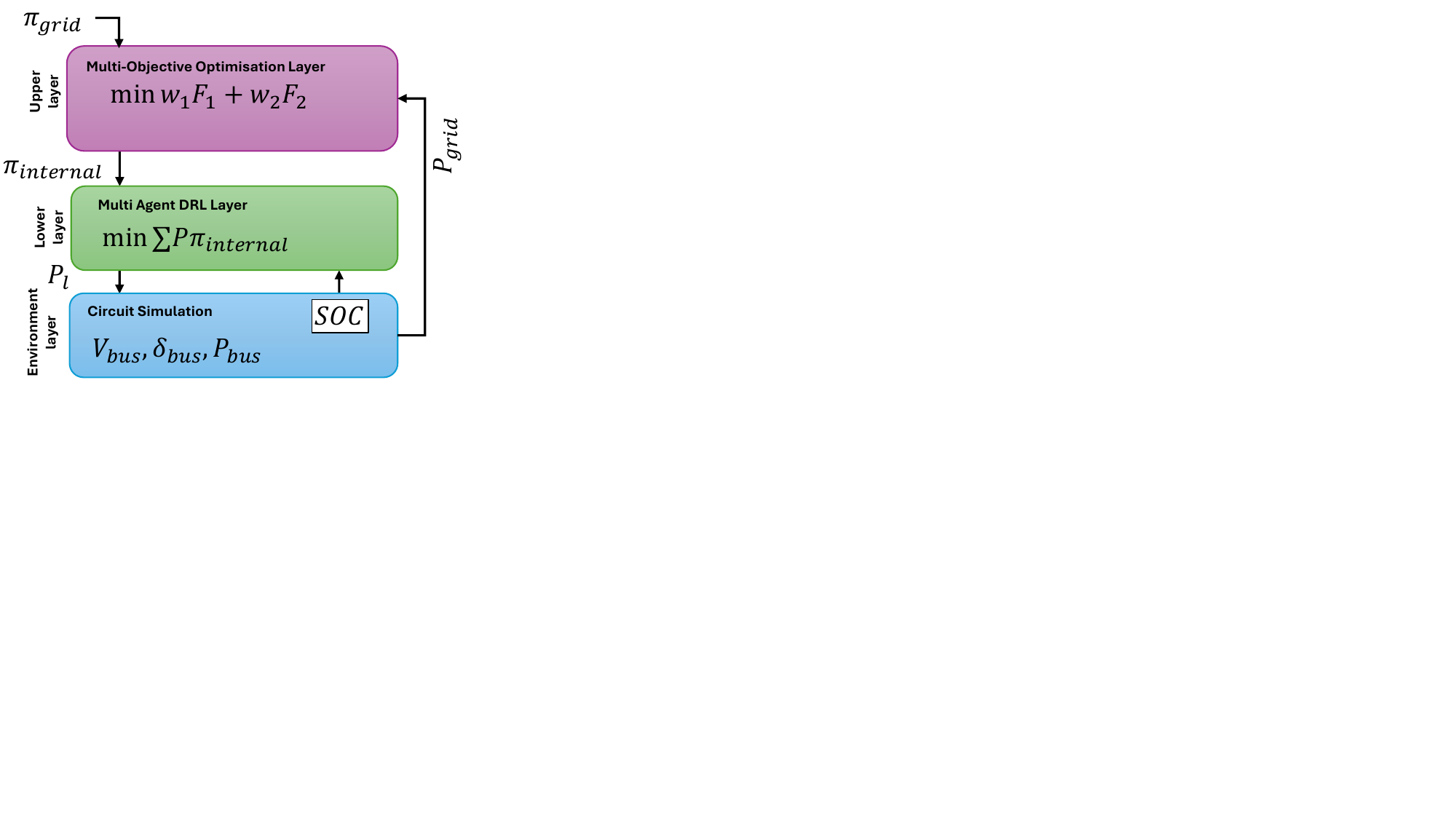}
    \put(2,73){a)}
    \end{overpic}
\end{subfigure}%
\end{minipage}%
\hfill
\begin{minipage}[c]{0.5\textwidth}
\centering
\begin{subfigure}{1\textwidth}
    \begin{overpic}[width=1\linewidth, trim = {0.0cm 4cm 19cm 0cm}, clip]{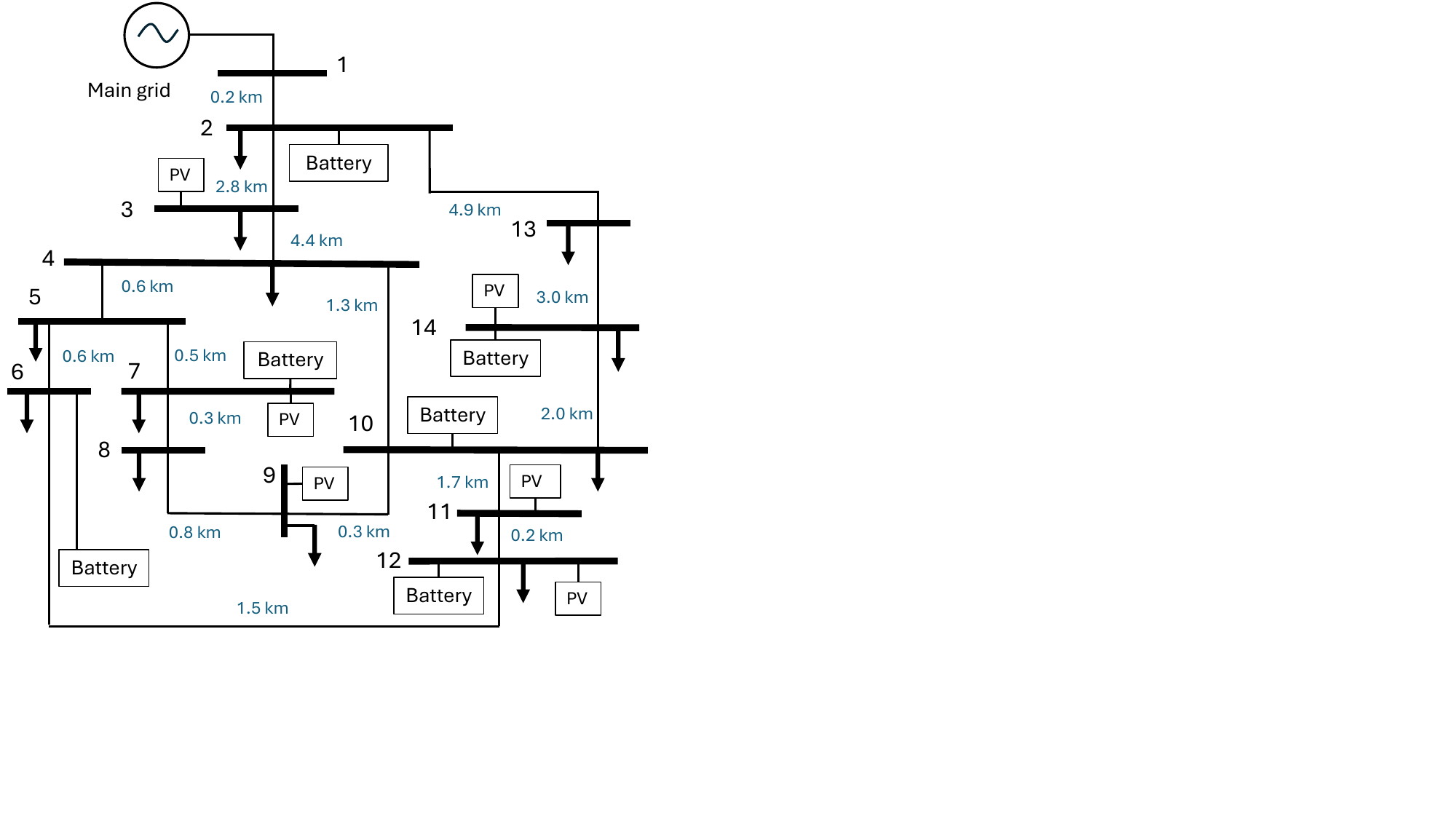}
    \put(3,81){b)}
    \end{overpic}
\end{subfigure}
\end{minipage}

\caption{a) Proposed control framework. b) Simulated distribution circuit model. Numbers in black indicate the Bus, the numbers in blue indicate line distance between buses and the arrows indicate electrical loads.}
\label{FOO}

\end{figure}

The proposed control system is divided into two layers. The lower layer performs local optimisation in the distribution system using a multi-agent system, where each agent controls a single battery based on the price of electricity provided by the upper layer. The upper layer adjusts the internal price of electricity to regulate the interaction with the main grid. These control layers form a distributed control \cite{r21} system as shown in Figure \ref{FOO}.

\subsection{Multi-Objective Layer}

The upper control layer performs dynamic pricing by solving the next MOO problem using heuristic optimisation, namely differential evolution:

\begin{mini!}|s|
{\alpha, \beta}{ w_{1} F_{1}(t) +  w_{2} F_{2}(t)  }
{}{}, \forall t \in T
\addConstraint{\alpha, \beta \in [0,1]}
\addConstraint{w_{1}+w_{2} = 1}
\end{mini!}

where $w_{1}$ and $w_{2}$ are weights for the objectives $F_{1}$ and $F_{2}$, at each time step $t$ for a duration of $T$. The first objective is the cost minimisation of the entire system and the second objective is the minimisation of power flow $P_{grid}$ with the main grid, expressed as:

\begin{equation}
F_{1} = c_{1}\sum P_{grid}(t)\pi_{grid}
\end{equation}

\begin{equation}
F_{2} =  c_{2}\sum (P_{grid}(t))^2
\end{equation}

where parameters $\alpha$ and $\beta$ modify the internal price of electricity using predictive and corrective measures of internal generation, and $c_1$ and $c_{2}$ normalise the objectives to one for $\alpha, \beta = 0$. The power flow $P_{grid}$ with the main grid depends on the line losses and power Load, $P_L$, internal Power generation $P_G$ and Power from batteries $P$:

\begin{equation}
P_{grid}(t) = P_L(t) - P_G(t) - P
\end{equation}

The predictive measure, defined as $\pi_{p}$ in $euro/kW$, is based on a forecast with length $K$ of the internal load and generation, $P_{Lf}$ and $P_{Gf}$ , calculated as follows:

\begin{equation}
    \pi_{p}(t+k) = (1-\alpha) \pi_{grid}(t+k) + (P_{Lf}(t+k) - P_{Gf}(t+k))\alpha, \forall k \in K
\end{equation}

The internal price $\pi_{internal}$ is further modified by a corrective measure to account for forecast errors and the reaction of other agents, based on $P_{grid}$ expressed in $euro/kW$, as follows:

\begin{equation}
    \pi_{internal}(t) = (1-\beta)\pi_{p}(t) +  \frac{P_{grid}(t)+P_{grid}(t-1 )}{2}\beta     
\end{equation}

This modifies the internal price according to the current net demand. To increase the price and grid stability, the price changes with a moving average model, using the current and previous price.

\subsection{Multi-Agent Layer}
The lower layer uses a distributed control system, with multiple independent DRL agents, to control each battery in the distribution circuit. In this paper, we use PPO for its ability to handle continuous action and state spaces and its stability during training \cite{r35, r36}. The DRL agent learns a policy to interact with its environment, in this case the circuit,  observing a state $S$ and taking an action $a$ sampled from $A$, to maximise the accumulated reward $R$.

The objective function $L$ of PPO during training is described as follows \cite{r35}:

\begin{equation}
    L(a) = \mathbb{E}_{t}[min( r_{t}(a) [R-R_{exp}],clip(r_{t}(a),1-\delta,1+\delta )[R-R_{exp}]]
\end{equation}

where $\mathbb{E}$ is the expectation function at time $t$; $r_{t}$ is the ratio of probabilities of the new policy over the old policy of selecting action $a$; $R_{exp}$ is the expected reward observing the state $S$ and $\delta$ is a hyperparameter that limits policy changes between updates. In this paper, the states observed by the DRL agent are expressed as follows:
\begin{equation}
S = \{P_{RB}, SOC, \pi_{internal} \}
\end{equation}
where $P_{RB}$ is a rule-based power schedule and $SOC$ is the battery state of charge. The rule-based battery power schedule charges at the lowest forecasted price and discharges at the highest. 

Based on the state observed at each hour, the agent selects a power reference $P$ for the battery sampled from the action space $A \in [-P_{max},P_{max}]$. This is used to define the power schedule of each battery. During training, the power schedule is used to calculate the agent's reward. In this case, the reward is the total cost of energy, that is:

\begin{mini!}|s|
{P}{     R = \sum P \pi_{internal}(t)  }
{}{}, \forall t \in T
\addConstraint{0 \leq SOC \leq SOC_{max}}
\end{mini!}

where $SOC_{max}$ is the maximum $SOC$. $SOC$ is calculated as follows with charge and discharge efficiencies, $\eta_{c}$ and $\eta_{d}$:

\begin{equation}\label{soc}
    SOC(t+1)= \left\{ \begin{array}{ll} SOC(t)+\eta_{c}P \quad P>0 \\ SOC(t)+\eta_{d}P \quad \text{else} \end{array} \right. 
\end{equation}

\begin{figure}[h!]
    \centering

\begin{subfigure}{0.5\textwidth}
    \begin{overpic}[width=1\textwidth, trim = {0.25cm 0 0.25cm 0.25cm}, clip]{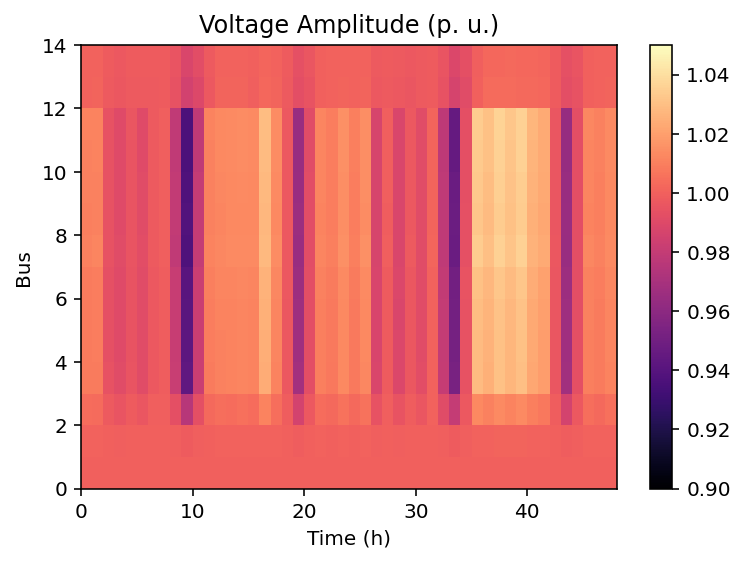}
     \put(10,75){a)}
    \end{overpic}
\end{subfigure}%
\hfill
\begin{subfigure}{0.5\textwidth}
    \begin{overpic}[width=1\textwidth, trim = {0.2cm 0 0.25cm 0.25cm}, clip]{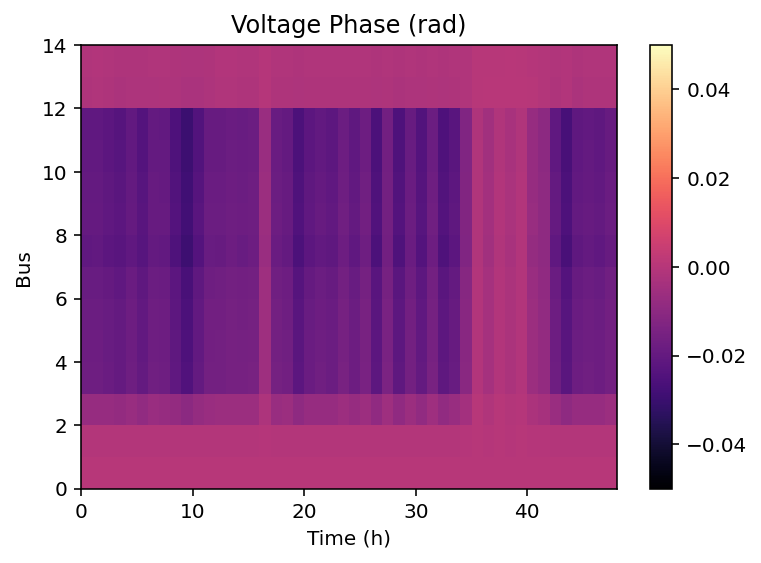}
    \put(10,74){b)}
    \end{overpic}
\end{subfigure}%
\\
\centering
\begin{subfigure}{0.45\textwidth}
    \begin{overpic}[width=1\textwidth, trim = {0.2cm 0 0.25cm 0.25cm}, clip]{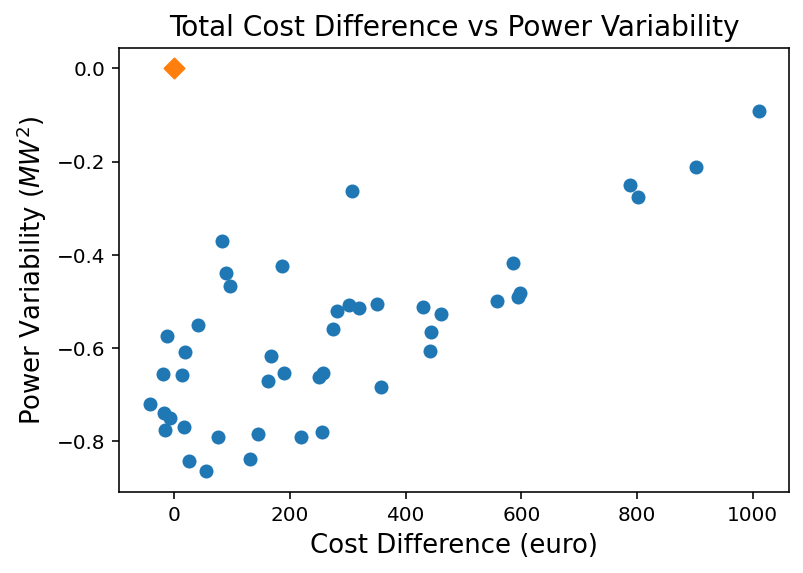}
    \put(10,70){c)}
    \end{overpic}
\end{subfigure}%
\hspace{0.005\textwidth}
\centering
\begin{subfigure}{0.45\textwidth}
    \begin{overpic}[width=1\textwidth, trim = {0.2cm 0 0.25cm 0.25cm}, clip]{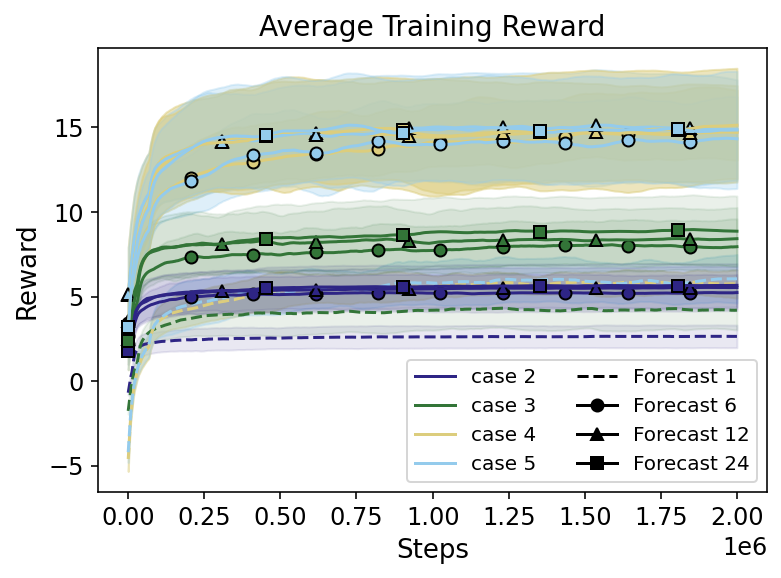}
    \put(10,72){d)}
    \end{overpic}
\end{subfigure}%

\begin{subfigure}{1\textwidth}
\centering
    \begin{overpic}[width=0.6\textwidth, trim = {0cm 0 0cm 0cm}, clip]{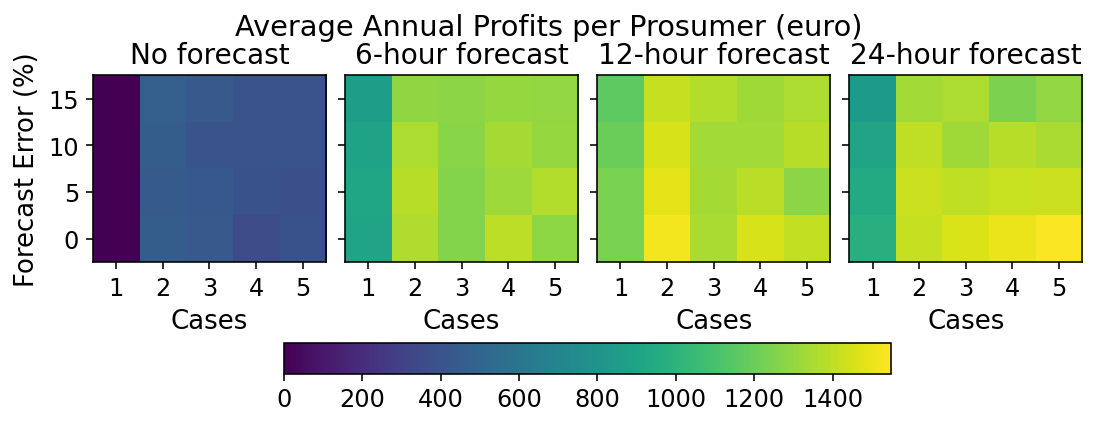}
    \put(-5,36){e)}
    \end{overpic}
\end{subfigure}%
\hfill

    \caption{Simulation results. a) Voltage amplitude response during peak PV generation in volts per unit (p. u.). b) Phase angle response during peak PV generation. c) MOO results for the distribution grid. The orange dot represents the results with no dynamic pricing. d) Training response of the DRL agent. e) Profits from energy arbitrage for a single prosumer in average.}
    \label{abcd}

\end{figure}

\section{Test Case}

The Multi-Objective layer implements differential evolution using Scipy with parameters specified in Appendix A, and the DRL agents are trained using PPO from stable baselines 3 (SB3). $w_1$ is varied between 0.1 and 0.9 in increments of 0.1 to test different weights of each objective, running each configuration 10 times using random seeds. We tested the Irish single market price data from 2022 publicly available from the Single Electricity Market Operator (SEMO), with equal buying and selling prices. The circuit simulation model is shown in Figure \ref{FOO} using PyPSA and the distributed generation and storage capacity connected to each Bus, that is, a voltage measurement point, is presented in Appendix B. $\eta_{c}$ and $\eta_{d}$ are set to $0.95$ and $1/0.95$, equivalent to a round-trip battery efficiency of 90\%, found in Tesla Powerwalls. The simulated load represents 50 Irish farms distributed across the circuit model \cite{r37}, and the simulated renewable generation is based on Ireland. The nominal voltage is $10KV$ and the lines have an impedance of $0.157+i0.123 \; \Omega/km$. The price forecast is provided to each DRL agent with lengths of 0 (no forecast), 6, 12 and 24 hours with 0, 5, 10 and 15\% added error to represent realistic conditions. PV generation and load forecast is provided similarly for dynamic pricing.

The DRL agent uses two fully connected layers, testing 5 cases: 1) using only the rule-based model. 2) using the DRL agent with 7-day trajectories for training with 64 neurons in each layer. 3) 14-day trajectories with 62 neurons in each layer. 4) 30-day trajectories with 128 neurons in each layer. 5) 30-day trajectories with 256 neurons in each layer. The training lasts 2 million steps, where each step represents one hour. In total, 640 configurations are tested.

\section{Simulation Results}

During peak renewable generation during 4th and 5th of May of 2022, the simulation results show that voltage and phase angle variations are in compliance with the limits of the Irish grid of 1 volt  per unit (p. u.) $\pm 10\%$ and phase angle of 0 $\pm 0.69$ radians , on a base of 230 Volts per phase, established in the EN 50160 standard. The simulated scenario represents the highest variation in the electrical response for the data tested and is shown in Figure \ref{abcd}.

In terms of energy arbitrage in each battery, results show highest profits using a 24-hour price forecast in case 5 with 1866 euros. In comparison, using only the rule-based model, the highest profit achieved is 1569 euros. The results using the DRL agents also show that errors of 5\% do not significantly affect total costs. Similarly, Figure \ref{abcd} shows that there is no significant difference between the 12-hour and 24-hour forecast. The total cost of energy for the entire system for one year is 55, 263 euros and the power variability is 34.73 $MW^2$.

The simulation results show that the Multi-Objective layer can reduce power variability of the entire system with the main grid by 0.85$MW^2$, without disrupting the cost minimisation achieved by the Multi-Agent layer, using dynamic pricing with parameters $\alpha, \beta = [0.022,0.305]$, as shown in Figure \ref{abcd}.

\section{Conclusion}
This paper presents a hierarchical and distributed energy management framework for the rural sector combining Deep Reinforcement Learning and differential evolution for Multi-Objective Optimisation. The use of DRL agents achieves up to 18\% increase in profits from energy arbitrage compared to using a rule-based power schedule. The use of dynamic pricing can reduce the power variation with the main grid by $0.85 MW^2$ and maintain a similar cost compared to using only the DRL agents, which promotes the use of distributed generation and voltage stability. In addition, the simulation of the voltage response shows that the proposed control system is in compliance with the Irish grid standard EN 50160. Furthermore, simulation results show that a relatively small difference in the accuracy of the electricity price forecast did not significantly affect profits from energy arbitrage.

\begin{credits}
\subsubsection{\ackname} This research was conducted with the financial support of the Science Foundation Ireland under Grant number [21/FFP-A/9040].
\end{credits}

\newpage

\printbibliography

@ARTICLE{r3,
  author={Li, Xiaoyu and Han, Xueshan and Yang, Ming},
  journal={IEEE Access}, 
  title={Day-Ahead Optimal Dispatch Strategy for Active Distribution Network Based on Improved Deep Reinforcement Learning}, 
  year={2022},
  volume={10},
  pages={9357-9370},
  keywords={Uncertainty;Optimization;Renewable energy sources;Reinforcement learning;Heuristic algorithms;Adaptation models;Active distribution networks;Active distribution network;multistage stochastic programming;optimal dispatch;data driven;deep learning;deep reinforcement learning},
  %doi={10.1109/ACCESS.2022.3141824}
}

@ARTICLE{r14,
  author={Chen, Jingwen and Mao, Lei and Liu, Yaoxian and Wang, Jinfeng and Sun, Xiaochen},
  journal={IEEE Access}, 
  title={Multi-Objective Optimization Scheduling of Active Distribution Network Considering Large-Scale Electric Vehicles Based on NSGAII-NDAX Algorithm}, 
  year={2023},
  volume={11},
  pages={97259-97273},
  keywords={Optimization;Scheduling;Electric vehicle charging;Monte Carlo methods;Load modeling;Distribution networks;Active distribution networks;Fuzzy systems;Electric vehicle;active distribution network;multi-objective optimization;NSGAII-NDAX algorithm;fuzzy theory},
  %doi={10.1109/ACCESS.2023.3312573}
  }

@ARTICLE{r15,
  author={Akbar, Muhammad Imran and Kazmi, Syed Ali Abbas and Alrumayh, Omar and Khan, Zafar A. and Altamimi, Abdullah and Malik, M. Mahad},
  journal={IEEE Access}, 
  title={A Novel Hybrid Optimization-Based Algorithm for the Single and Multi-Objective Achievement With Optimal DG Allocations in Distribution Networks}, 
  year={2022},
  volume={10},
  pages={25669-25687},
  keywords={Voltage;Optimization;Resource management;Stability criteria;Genetic algorithms;Distribution networks;Costs;Distributed generation;dimension learning-based hunting;grey wolf optimization;particle swarm optimization;radial distribution network;voltage deviation;voltage stability index},
  %doi={10.1109/ACCESS.2022.3155484}
  }

@ARTICLE{r16,
  author={Xu, Ruipeng and Zhang, Cuo and Xu, Yan and Dong, Zhaoyang and Zhang, Rui},
  journal={IEEE Transactions on Smart Grid}, 
  title={Multi-Objective Hierarchically-Coordinated Volt/Var Control for Active Distribution Networks With Droop-Controlled PV Inverters}, 
  year={2022},
  volume={13},
  number={2},
  pages={998-1011},
  keywords={Inverters;Reactive power;Voltage control;Uncertainty;Real-time systems;Power generation;Stochastic processes;Active distribution network;droop control;photovoltaic;volt/Var control;uncertainty},
  %doi={10.1109/TSG.2021.3126761}
  }

@ARTICLE{r19,
  author={Sarantakos, Ilias and Peker, Meltem and Zografou-Barredo, Natalia-Maria and Deakin, Matthew and Patsios, Charalampos and Sayfutdinov, Timur and Taylor, Phil C. and Greenwood, David},
  journal={IEEE Transactions on Smart Grid}, 
  title={A Robust Mixed-Integer Convex Model for Optimal Scheduling of Integrated Energy Storage—Soft Open Point Devices}, 
  year={2022},
  volume={13},
  number={5},
  pages={4072-4087},
  keywords={Uncertainty;Schedules;Optimal scheduling;Load flow;Costs;Batteries;DC-DC power converters;Converter losses;convex optimization;energy storage;robust optimization;soft open~point},
  %doi={10.1109/TSG.2022.3145709}
  }

@ARTICLE{r20,
  author={Zhang, Zhijun and Zhang, Yudi and Yue, Dong and Dou, Chunxia and Ding, Xiaohua and Zhang, Huifeng},
  journal={IEEE Transactions on Industrial Informatics}, 
  title={Economic-Driven Hierarchical Voltage Regulation of Incremental Distribution Networks: A Cloud-Edge Collaboration Based Perspective}, 
  year={2022},
  volume={18},
  number={3},
  pages={1746-1757},
  keywords={Voltage control;Economics;Collaboration;Security;Optimization;Delay effects;Uncertainty;Cloud-edge collaboration;economic operation;incremental distribution networks;model predictive control;voltage regulation},
  %doi={10.1109/TII.2021.3085670}
  }

@ARTICLE{r21,
  author={Wu, Jiang and Wu, Longkun and Xu, Zhanbo and Qiao, Xiaoyi and Guan, Xiaohong},
  journal={IEEE Transactions on Automation Science and Engineering}, 
  title={Dynamic Pricing and Prices Spike Detection for Industrial Park With Coupled Electricity and Thermal Demand}, 
  year={2022},
%  volume={19},
%  number={3},
  pages={1326-1337},
  keywords={Pricing;Demand response;Power system dynamics;Load modeling;Costs;Predictive models;Mathematical models;Industrial park;dynamic pricing;demand response;Lagrangian relaxation;LSTM},
  %doi={10.1109/TASE.2021.3139825}
  }

@ARTICLE{r28,
  author={Wang, Yi and Qiu, Dawei and Strbac, Goran and Gao, Zhiwei},
  journal={IEEE Transactions on Industrial Informatics}, 
  title={Coordinated Electric Vehicle Active and Reactive Power Control for Active Distribution Networks}, 
  year={2023},
  volume={19},
  number={2},
  pages={1611-1622},
  keywords={Reactive power;Voltage control;Training;Uncertainty;Reactive power control;Privacy;Decentralized control;Active and reactive power control;active distribution networks (ADNs);electric vehicles (EVs);location marginal prices (LMPs);multiagent reinforcement learning (MARL)},
  %doi={10.1109/TII.2022.3169975}
  }

@article{r32,
title = {Multi objectives reinforcement learning for smart buildings: A systematic review of algorithms, applications and future perspectives},
journal = {Energy and Buildings},
volume = {345},
pages = {116045},
year = {2025},
%issn = {0378-7788},
%doi = {https://doi.org/10.1016/j.enbuild.2025.116045},
%url = {https://www.sciencedirect.com/science/article/pii/S0378778825007753},
author = {Thi Ngoc Yen Huynh and Anh Tuan Nguyen and Yonghan Ahn and Bee Lan Oo and Benson T.H. Lim},
keywords = {Multi-objectives reinforcement learning, Multi-objectives optimization, Smart building, Systematic review, Decision-making strategies},
}

@misc{r35,
      title={Proximal Policy Optimization Algorithms}, 
      author={John Schulman and Filip Wolski and Prafulla Dhariwal and Alec Radford and Oleg Klimov},
      year={2017},
      eprint={1707.06347},
      archivePrefix={arXiv},
      primaryClass={cs.LG},
      url={https://arxiv.org/abs/1707.06347}, 
}

@InProceedings{r36,
author="Ali, Nawazish
and Shaw, Rachael
and Mason, Karl",
editor="Mathieu, Philippe
and De la Prieta, Fernando",
title="A Deep Reinforcement Learning Approach to Battery Management in Dairy Farming via Proximal Policy Optimization",
booktitle="Advances in Practical Applications of Agents, Multi-Agent Systems, and Digital Twins: The PAAMS Collection",
year="2025",
publisher="Springer Nature Switzerland",
address="Cham",
pages="15--26",
%abstract="Dairy farms consume a significant amount of electricity for their operations, and this research focuses on enhancing energy efficiency and minimizing the impact on the environment in the sector by maximizing the utilization of renewable energy sources. This research investigates the application of Proximal Policy Optimization (PPO), a deep reinforcement learning algorithm (DRL), to enhance dairy farming battery management. We evaluate the algorithm's effectiveness based on its ability to reduce reliance on the electricity grid, highlighting the potential of DRL to enhance energy management in dairy farming. Using real-world data our results demonstrate how the PPO approach outperforms Q-learning by 1.62{\%} for reducing electricity import from the grid. This significant improvement highlights the potential of the Deep Reinforcement Learning algorithm for improving energy efficiency and sustainability in dairy farms.",
isbn="978-3-031-70415-4"
}

@INPROCEEDINGS{r37,
  author={Khaleghy, Hossein and Clifford, Eoghan and Mason, Karl},
  booktitle={2024 IEEE 6th International Conference on Cybernetics, Cognition and Machine Learning Applications (ICCCMLA)}, 
  title={A Machine Learning Approach to Dairy Farm Energy Disaggregation}, 
  year={2024},
  volume={},
  number={},
  pages={55-60},
  keywords={Measurement;Machine learning algorithms;Electricity;Computational modeling;Machine learning;Data models;Solar panels;Monitoring;Load modeling;Synthetic data;Machine Learning;Energy;Energy Disaggregation},
%  doi={10.1109/ICCCMLA63077.2024.10871399}
  }

\newpage

\section*{Appendix A}
The parameters for differential evolution used in the Multi-Objective Layer are described in Table \ref{t3}.

\begin{table}[]
    \centering
    \begin{tabular}{c|c}
         Parameter& Value \\
         \hline
   Population      &  30\\
   Mutation & [0.5,1]\\
   Convergence Tolerance & 0.01 \\
   Maximum number of iterations & 1000\\
   Crossover rate = 0.7\\
   
    \end{tabular}
    \caption{Differential Evolution Parameters}
    \label{t3}
\end{table}

\section*{Appendix B}
The distributed generation and storage capacity in the circuit model shown in Figure \ref{FOO} is described in Table \ref{t2}.

\begin{table}[h!]
    \centering
    \begin{tabular}{c|cc| cc}
    & \multicolumn{2}{|c|}{Battery}&\multicolumn{2}{c}{Solar PV}\\ 
    \hline
         Bus& $P_{max}$ (kW)& $SOC_{max}$ (kWh) & Bus& $P_{max}$ (kW)  \\

              \hline \hline
            2& 20 &27  &3& 22.44\\
            6&30&40.5 &7&39.27 \\
            7&50&67.5 &9& 33.66\\
            10& 40&54  &11&28.05 \\
            12& 40&54  &12&28.05 \\
            14&10&13.5 & 14& 16.83\\
    \end{tabular}
    \caption{Distributed generation specifications by Bus. The left side describes the parameters for the batteries and the right side describes the PV generation.}
    \label{t2}
\end{table}

\end{document}